\documentclass[PHME, 2024]{PHMSociety}

\usepackage{graphicx}
\usepackage{amsmath}
\usepackage{enumitem}
\usepackage{booktabs}
\usepackage{siunitx}
\usepackage{multirow}
\usepackage[colorlinks,citecolor=orange,
    linkcolor=cyan, 
    filecolor=cyan, 
    urlcolor=cyan]{hyperref}
\usepackage{cleveref} 
\usepackage{apacite} 

\begin{document}

\title{Virtual Sensor for Real-Time Bearing Load Prediction Using Heterogeneous Temporal Graph Neural Networks}

\author{%
	Mengjie Zhao\authorNumber{1}, Cees Taal\authorNumber{2}, Stephan Baggerohr \authorNumber{3}, and Olga Fink\authorNumber{4}
}

\address{
	\affiliation{{1,4}}{Intelligent Maintenance and Operations Systems, EPFL, Lausanne, Switzerland}{ 
		{\email{mengjie.zhao@epfl.ch}}\\ 
		{\email{olga.fink@epfl.ch}}
		} 
	\tabularnewline 
	\affiliation{2, 3}{SKF, Research and Technology Development, Houten, the Netherlands}{ 
		{\email{cees.taal@skf.com}}\\
            {\email{stephan.baggerohr@skf.com}}
		} 
}

\maketitle
\pagestyle{fancy}
\thispagestyle{plain}

\phmLicenseFootnote{Mengjie Zhao}

\begin{abstract}%
Accurate bearing load monitoring is essential for their Prognostics and Health Management (PHM), enabling damage assessment, wear prediction, and proactive maintenance. 
While bearing sensors are typically placed on the bearing housing, direct load monitoring requires sensors inside the bearing itself. 
Recently introduced sensor rollers enable direct bearing load monitoring but are constrained by their battery life. 
Data-driven virtual sensors can learn from sensor roller data collected during a battery's lifetime to map operating conditions to bearing loads.
Although spatially distributed bearing sensors offer insights into load distribution (e.g., correlating temperature with load), traditional machine learning algorithms struggle to fully exploit these spatial-temporal dependencies. 
To address this gap, we introduce a graph-based virtual sensor that leverages Graph Neural Networks (GNNs) to analyze spatial-temporal dependencies among sensor signals, mapping existing measurements (temperature, vibration) to bearing loads.
Since temperature and vibration signals exhibit vastly different dynamics, we propose Heterogeneous Temporal Graph Neural Networks (HTGNN), which explicitly models these signal types and their interactions for effective load prediction. 
Our results demonstrate that HTGNN outperforms Convolutional Neural Networks (CNNs), which struggle to capture both spatial and heterogeneous signal characteristics. 
These findings highlight the importance of capturing the complex spatial interactions between temperature, vibration, and load.

\end{abstract}

\section{Introduction}
Bearings are essential components in mechanical systems, ensuring the efficient and reliable operation of machinery and equipment across diverse industries, including wind energy, aerospace, and automotive sectors. 
Real-time monitoring of bearing conditions is crucial for optimal performance and proactive maintenance~\shortcite{hou2021measurement}. 
Knowing the actual load experienced by bearings offers several key benefits.
Firstly, deviations from the original design loads signal the need for adjustments to operational parameters and maintenance schedules. This allows for proactive prescription of health-aware load profiles, potentially extending the bearing's service life. Moreover, load monitoring aids in early detection of misalignments, enabling timely proactive adjustments to prevent further damage~\shortcite{widner1976bearing}. 
Additionally, knowledge of bearing load facilitates more accurate diagnosis of potential bearing faults~\cite{peng2020multibranch}. 
Finally, bearing loads are a key factor influencing bearing lifespan and failure \shortcite{harris2006}, and their understanding enables predicting damage propagation \shortcite{morales2019}. An in-depth understanding of the load is essential for accurate Remaining Useful Life (RUL) prediction and effective Prognostic and Health Management (PHM).


Directly measuring bearing loads during operation presents complex challenges. Traditional approaches, typically using strain gauges, require direct contact or close proximity to the bearing's rolling elements. This introduces significant logistical and technical hurdles \shortcite{konopka2023}, including accessing power and establishing sensor communication, making installation more expensive than conventional condition monitoring sensors, such as for vibration and temperature.  

Recently, wireless sensor roller technology has been introduced, wherein sensors are embedded inside a rolling element to allow in-operation measurement of bearing loads \cite{baggerohr2023}. 
However, their utility is still constrained by battery life. A virtual sensor could overcome this limitation by providing continuous, long-term load predictions, even when the sensor roller's battery is depleted. 
Specifically, our goal is to develop a virtual sensor that maps the measurements of conventional bearing condition monitoring sensors to loads.
Since the relationships between these sensors and load are influenced by factors such as stiffness, damping, and thermal behavior, and are often unknown in real-world applications, we adopt a data-driven approach.
Sensor roller provides crucial ground-truth load data, which is significant for enabling the development of this virtual sensor. Estimating the load without such direct data is extremely difficult without extensive modeling. 
Our approach not only extends the value of the physical sensor roller but also supports advanced PHM.

Virtual sensors have been applied in many different applications ranging from environmental sensing to complex industrial systems. They leverage readily available measurements and computational models to infer quantities that are challenging or costly to measure directly \shortcite{martin2021virtual}. 
They also play a crucial role in digital twins, providing insights beyond what physical sensors can capture \cite{song2023digital}. 
Two primary directions exist for virtual sensors: \textit{model-based} and \textit{data-driven}. Model-based approaches rely on well-defined physical laws and principles to develop models describing the system of interest. In contrast, \textit{data-driven} approaches use machine learning and data mining algorithms to find patterns and relationships within sensor data.
Model-based virtual sensors require using existing sensor data to accurately infer and update model parameters to ensure accurate estimations.
Methods such as Kalman filtering, which dynamically updates model states in real-time based on noisy sensor measurements, are well-established for calibrating physics-based virtual sensors for load estimation models~\shortcite{kerst2019model}.
Alternatively, Gaussian processes can be applied to latent force models to infer unknown load dynamics from a sensor network~\shortcite{bilbao2022virtual}. 
While powerful, these methods rely on prior knowledge of the system's physics, which can be challenging or infeasible to obtain in many real-world cases.
In contrast, data-driven virtual sensing offers flexibility by directly learning complex relationships from data. For example, \shortcite{dimitrov2022virtual} demonstrated the potential of Long Short-Term Memory (LSTM) networks for predicting wind turbine blade root bending moment using SCADA data. 
\shortcite{wang2021soft} developed a Deep Belief Network (DBN) with event-triggered learning (DBN-EL) to improve the efficiency and accuracy of a water quality soft-sensing model for the wastewater treatment processes from the sensor data.

Model-based methods often depend on prior knowledge of the system's physics. In contrast, data-driven approaches can overcome this limitation but may require other forms of ground truth to learn the functional relationships, such as simulation data~\cite{dimitrov2022virtual} or periodic lab-based measurements~\cite{wang2021soft}, making them difficult to apply in real-world scenarios.
Fortunately, bearing sensor rollers allow direct measurement of bearing load in operation, offering a direct ground truth that enables us to learn the complex relationships between load and conventional bearing condition monitoring sensors through supervised learning.

For large-size bearings, such as main bearings in wind turbines, a common approach for bearing condition monitoring involves positioning multiple sensors around the bearing to measure rotational speed, vibration, and temperature. 
Although a correlation exists between load and these sensor readings, the relationships are complex and difficult to model accurately due to the lack of exact physical models.
However, there exists an additional inductive bias in the form of spatial information, such as the correlation between higher temperatures and areas of increased load.
Leveraging this spatial information can offer valuable insights into load distribution.
While traditional machine learning algorithms struggle to effectively utilize this spatial information, Graph Neural Networks (GNNs) are well-suited for handling spatial-temporal dependencies~\cite{jin2023survey}.
By modeling sensors and their connections as a graph, GNNs can directly capture the spatial dependencies and relationships between different sensor readings. They utilize message-passing techniques, where information from neighboring sensors is iteratively processed and aggregated, building a global understanding from local information~\shortcite{gilmer2017neural}. 
GNNs have been successfully applied in areas such as bearing remaining useful life prediction~\shortcite{yang2022bearing}, cyber-physical attack detection for water distribution systems~\cite{deng2021graph}, sensor calibration for air pollution~\shortcite{niresi2023spatial} and fault detection for chemical process plants~\shortcite{zhao2023dyedgegat}. 

Nevertheless, existing GNN methods often assume relatively similar feature characteristics across nodes. 
Although GNNs have been applied to heterogeneous sensor networks, the focus has typically been on handling different sensor types (e.g., temperature, humidity, pressure). In these cases, while the data originates from diverse sources, the signal characteristics often exhibit some similarities.
In our scenario, the heterogeneity is in signal characteristics. Vibration and temperature signals exhibit very different dynamics and frequencies. This poses a novel and significant challenge for GNNs, which often struggle to effectively integrate and learn from such highly diverse signal characteristics.

To address the challenge of heterogeneous sensor characteristics, we propose a novel virtual load sensor based on Heterogeneous Temporal Graph Neural Networks (HTGNNs).  
By explicitly modeling high and low-frequency signals as distinct node types and differentiating their interaction types, our HTGNN effectively fuses the information from diverse sensors. 
This enables more accurate load prediction, overcoming the limitations of traditional GNNs. 
To the best of our knowledge, this represents the first design of such an architecture to analyze diverse sensor types for bearing load estimation.

The remainder of this paper is organized as follows: Sec.~\ref{sec:problem_statement} describes the task of a bearing virtual sensor.
Sec.~\ref{sec:framework} elaborates on HTGNN’s core components to model the heterogeneous dynamic relationships within the bearing system. Sec.~\ref{sec:casestudy} describes the case study, experimental setup, and the baseline method Sec.~\ref{sec:result} presents the results of and offers a thorough discussion. Finally, Sec.~\ref{sec:conclusion} summarizes key findings and proposes directions for further research.

\section{Virtual Sensor for Load Prediction}
\label{sec:problem_statement}
In this paper, we establish the notation where bold uppercase letters (e.g., $\mathbf{X}$), bold lowercase letters (e.g., $\mathbf{x}$), and calligraphic letters (e.g., $\mathcal{V}$) to denote matrices, vectors, and sets, respectively. Time steps are indicated by Superscripts (e.g., $\mathbf{X}^t$ is the matrix $\mathbf{X}$ at time $t$), while subscripts identify specific nodes (e.g., $\mathbf{x}_i$ is the vector for node $i$).

\subsection{Problem Statement}
In our case study, we focus on monitoring a bearing with a heterogeneous network of sensors. The data are collected from a test rig and comprise $N$ sensor signals captured at discrete time instances. 
We particularly examine temperature and vibration data, which are represented as vectors:
\begin{align}
    \mathbf{x_T}^{t} &= [x_{T_1}^{t}, x_{T_2}^{t}, ..., x_{T_{N_T}}^{t}]^T \in \mathbb{R}^{N_T}, \\
    \mathbf{x_V}^{t} &= [x_{V_1}^{t}, x_{V_2}^{t}, ..., x_{V_{N_V}}^{t}]^T \in \mathbb{R}^{N_V},
\end{align}
where $N_T$ and $N_V$ are the number of each sensor type, while $x_{T_i}^{t}$ and $x_{V_i}^{t}$ denote the measurements at time $t$ from the $i^{th}$ sensor for temperature and vibration. 
Additionally, the rotational speed is recorded as $w^{t} \in \mathbb{R}$ at time $t$. Importantly, this characterizes the system's operational state and acts as a control parameter, rather than being a direct sensor measurement.
To construct time-series samples for each sensor type, we employ a sliding window of length $L$, resulting in the following representations:
\begin{align}
\mathbf{X_T}^{t_l:t} &= [\mathbf{x_T}^{t_l}, \cdots, \mathbf{x_T}^{t-1}, \mathbf{x_T}^{t}] \in \mathbb{R}^{N_T \times L},\\
\mathbf{X_V}^{t_l:t} &= [\mathbf{x_V}^{t_l}, \cdots, \mathbf{x_V}^{t-1}, \mathbf{x_V}^{t}] \in \mathbb{R}^{N_V \times L},\\
\mathbf{w}^{t_l:t} &= [{w}^{t_l}, \cdots, {w}^{t-1}, {w}^{t}] \in \mathbb{R}^{L},
\end{align}
where $t_l = t - L + 1 > 0$ marks the beginning of the observation window.

Our goal is to develop a function $f$, referred to as a virtual sensor, to accurately estimate the bearing load $\mathbf{y}^t \in \R^d$ at time $t$, targeting both axial and radial loads ($d=2$). This function learns from heterogeneous sensor data $\mathbf{X_T}^{t_l:t}$, $\mathbf{X_V}^{t_l:t}$, and $\mathbf{W}^{t_l:t}$.
Several challenges arise in developing such a function.
Firstly, temperature and vibration signals exhibit inherently distinct characteristics. 
Temperature signals, typically monitored at lower frequencies, reflect gradual changes in the system's thermal state. 
In contrast, vibration signals are captured at high frequencies, offering insights into the immediate mechanical interactions and anomalies within the system. 
These differences in frequency not only affect the data processing strategy but also the interpretation of these signals in real-time monitoring. Additionally, the dynamic operating conditions introduce further complexity. Variations in load, speed, and environmental factors can significantly alter the base characteristics of both temperature and vibration data.
\section{Graph-based Load Prediction Model}
\label{sec:framework}
\subsection{Framework Overview}
\begin{figure*}[tbhp]
  \centering
    \includegraphics[width=0.8\linewidth]{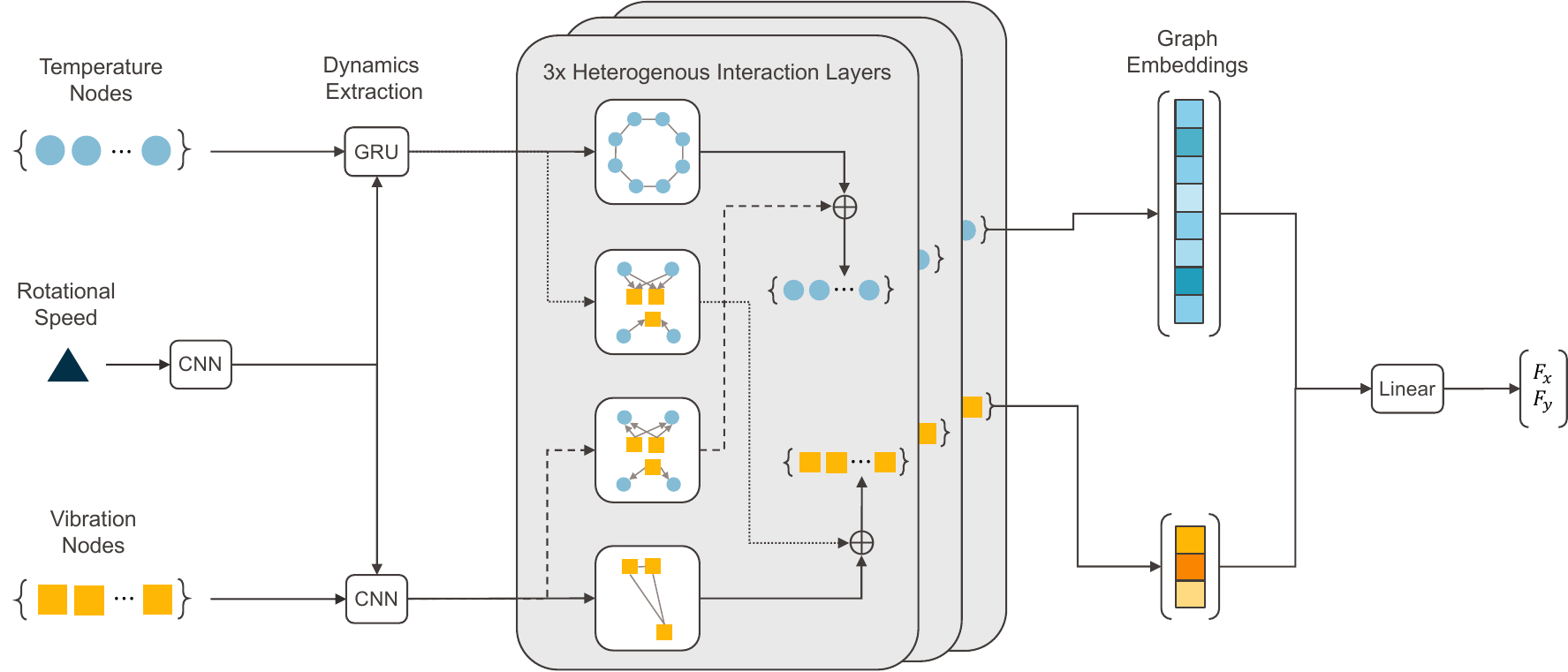}
  \caption{Architecture of the proposed Heterogeneous Temporal Graph Neural Network (HTGNN) for Load Prediction.}
  \label{fig:htgnn}
\end{figure*}
We propose a novel Heterogeneous Temporal Graph Neural Network (HTGNN) for real-time bearing load prediction. Our framework learns a virtual sensor function, $f(\mathbf{X_T}^{t_l:t}, \\\mathbf{X_V}^{t_l:t}, \mathbf{W_T}^{t_l:t})= \mathbf{Y}^{t}$, to accurately estimate the bearing load $\mathbf{Y}_t$ at a given time $t$.  
The HTGNN's main novelty lies in its ability to effectively capture the heterogeneity of sensor data and model the interactions between different sensor types.
We achieve this by representing different sensor types as distinct node types in an aggregated temporal graph. 
This allows us to extract unique dynamics of each sensor type using tailored models and then model their interactions with specialized GNNs, This offers a significant advantage over traditional homogeneous temporal GNN methods that consider only a single type of relation.
Fig.~\ref{fig:htgnn} illustrates the HTGNN architecture. The model's key components are:
\begin{enumerate}[
noitemsep,topsep=0pt,parsep=0pt,partopsep=0pt]
    \item \textbf{Heterogeneous temporal graph construction}, which constructs the bearing graph.
    (Sec.~\ref{sec:heteo_graph}).
    \item \textbf{Context-aware heterogeneous dynamics extraction}, \\which captures dynamics of different sensor types (Sec.~\ref{sec:htgnn_node_dynamcics}).
    \item \textbf{Heterogeneous interaction modelling}, which models complex interactions between diverse sensors (Sec.~\ref{sec:htgnn_interaction}).
    \item \textbf{Load prediction}, which predicts the bearing loads using the learned node representations (Sec.~\ref{sec:htgnn_load_prediction}).
\end{enumerate}
In the following, we detail each component of HTGNN.

\subsection{Heterogeneous Temporal Graph Construction}
\label{sec:heteo_graph}
\textbf{Heterogeneous Static Graph.} Following \cite{shi2022heterogeneous}, a Heterogeneous Static Graph (HSG), denoted as $\mathcal{G}=(\mathcal{V},\mathcal{E})$, consists of a node set $\mathcal{V}$ and an edge set $\mathcal{E}$, where nodes and edges can be of different types. The graph is associated with a node-type mapping function $\phi: \mathcal{V} \rightarrow \mathcal{A}$ and an edge-type mapping function $\psi: \mathcal{E} \rightarrow \mathcal{R}$, with $\mathcal{A}$ and $\mathcal{R}$ representing the sets of node and edge types, respectively, satisfying $|\mathcal{A}| + |\mathcal{R}| > 2$.

\textbf{Heterogeneous Temporal Graph.} Extending the concept of a Heterogeneous Static Graph (HSG), a Heterogeneous Temporal Graph (HTG) is defined as a sequence of HSGs over $T$ time steps, $\mathcal{G}^{T} = \{\mathcal{G}^{t_1}, \ldots, \mathcal{G}^{t_T}\}$. Each graph $\mathcal{G}^t = (\mathcal{V}^t, \mathcal{E}^t)$ within this sequence represents the state of the graph at time $t$. The node and edge type mapping functions, $\phi$ and $\psi$ remain consistent across time steps. The HTG can then be presented in an aggregated form as:
\begin{equation}
\mathcal{G}^{T}= \left(\bigcup_{t=t_1}^{t_T} \mathcal{V}^t, \bigcup_{t=t_1}^{t_T} \mathcal{E}^t\right),
\end{equation}
combining nodes $\mathcal{V}^t$ and edges $\mathcal{E}^t$ across all time steps while preserving heterogeneity defined by $\phi$ and $\psi$.

\textbf{Bearing graph construction.} To model the heterogeneous sensor signals from a sensor network of the bearing system, we construct an HTG. This graph consists of two types of nodes: temperature (T) with attributes $\mathbf{X_T}^{t_l:t}$ and vibration (V) with attributes $\mathbf{X_V}^{t_l:t}$. Edge types represent relationships between node types: T-T, V-V, T-V, and V-T. We assume that these relationships are invariant over time. The HTG allows capturing the interactions and evolution of temperature and vibration signals within the bearing system. A visualization of the HTG is provided in Fig.~\ref{fig:htgnn}.

\subsection{Context-aware Node Dynamics Extraction}
\label{sec:htgnn_node_dynamcics}
In complex systems, the behavior of individual nodes (sensors) is often influenced by the global operating context. In our bearing system, rotational speed can be considered a control variable, where increases in rotational speed lead to higher vibration intensity and faster temperature rises.
To capture these important influences, our HTGNN model leverages context-aware dynamics extraction for node, following the strategy proposed in~\cite{zhao2023dyedgegat}. 
We extract contextual information from rotational speed and integrate it into the dynamics modeling of other sensor types using tailored techniques.

\textbf{Rotational speed.}
To extract meaningful representations of operational state context from the rotational speed signal, which contains noise, we employ a 1D Convolutional Neural Network (1DCNN). We choose a 1DCNN due to its effectiveness in capturing patterns within time-series data.
This process generates a hidden representation of dimensionality $\mathbf{h}_w \in \R^{d_w}$, which is used to augment the dynamics extraction from other sensor types. Our 1DCNN configuration adopts  channel sizes [2, 2, 1], kernel sizes [3, 5, 5], and employs the SiLU activation function:
\begin{equation}
\mathbf{h}_w = \text{SiLU} \left(\text{1DCNN}(\mathbf{w}^{t_l:t})\right),
\label{eq:rot_cnn}
\end{equation}

\textbf{Temperature.} We model the temperature dynamics using a Gated Recurrent Unit (GRU) network. For each temperature node $j$, the GRU updates its cell state at each time step $\tau$ to capture the temporal dynamics within the sequence $\mathbf{x_T}_j^{t_l:t}$. Importantly, we initialize the GRU's hidden state with $\mathbf{h}_w$ (rotational speed encoding from Eq.~\ref{eq:rot_cnn}), allowing the operational state context to influence temperature dynamics:
\begin{equation}
    {\mathbf{h_{T}}}_i^{\tau} = \text{SiLU}\left(\text{GRU-Cell}(\mathbf{x_T}_i^{\tau}, \mathbf{h_T}_i^{\tau-1})\right), \forall \tau \in [t_l, t].
    \label{eq:temp_node_gru}
\end{equation}
We use the final state $\mathbf{h_T}_i^t \in \mathbb{R}^{d_T}$,  representing the encoded dynamics of node $i$ up to time $t$ and incorporating the operational state context, as the temperature node representation $\mathbf{h_T}_i \in \mathbb{R}^{d_T}$.

\textbf{Vibration.} Similar to the rotational speed encoding, we use a 1DCNN to model the dynamics of vibration signals. This process learns the hidden representation $\mathbf{h_V}_i^t$ from the vibration sequence $\mathbf{x_V}_i^{t_w:t}$ of a vibration signal $i$: 
\begin{equation}
\mathbf{h_V}_i^t = \text{SiLU} \left(\text{1DCNN}(\mathbf{x_V}_i^{t_l:t})\right).
\label{eq:vib_node_cnn}
\end{equation}
Finally, we concatenate $\mathbf{h_V}_i^t \in \mathbb{R}^{d_V}$ with $\mathbf{h_w}\in \R^{d_w}$ to form the complete node representation $\mathbf{h_V}_i = \left[\mathbf{h_V}_i^t \parallel \mathbf{h_w}\right] \in \R^{d_V+d_w}$. This incorporates both vibration dynamics and operational state.

\subsection{Heterogeneous Interaction Modelling}
\label{sec:htgnn_interaction}
We model heterogeneous interactions between different sensor types to capture the influence of operating context-aware dynamics. The proposed HTGNN model addresses two types of interactions within the graph:  interactions among the same type of nodes and interactions across different types. This interaction modeling applies to node dynamics previously extracted in the node dynamics extraction section (temperature node from Eq.~\ref{eq:temp_node_gru}, vibration node from Eq.~\ref{eq:vib_node_cnn}).

\textbf{Same-type interactions.} 
For interactions among nodes of the same sensor type, we employ Graph Convolutional Networks (GCNs)~\shortcite{kipf2017gcn}. This allows us to refine node representations by aggregating information from neighboring nodes that share similar characteristics.  
Messages passed from node $j$ to node $i$ of the same type with relation $r_s \in \mathcal{R}_{\text{s}}$ are computed as follows:
\begin{equation}
m_{j \rightarrow i}^{(l, r_s)} = \frac{1}{\sqrt{\hat{d}_i} \sqrt{\hat{d}_j}} \mathbf{W}_{\phi(j),r_s}^{(l)} \mathbf{h}_j^{(l)}, \forall r_s \in \mathcal{R}_{\text{s}}, \phi(j) = \phi(k),
\end{equation}
where $\hat{d}_i$ and  $\hat{d}_j$ denote normalized node degrees, and $\mathcal{R}_{\text{s}}$ is the set of edge types connecting nodes of the same type.

\textbf{Different-type interactions.} 
To model the influence of one sensor type on another (e.g., the impact of temperature on vibration), we utilize Graph Attention Networks v2 (GATv2) \shortcite{brody2022gatv2}. 
This mechanism dynamically computes attention-weighted messages, allowing the model to discern the varying importance of different neighbors. The attention coefficients $\alpha_{jk}^{(l,r_d)}$ for a target node $i$ receiving a message from node $j$ with relation $r_d \in\mathcal{R}_d$ are defined as:
\begin{equation}
\alpha_{jk}^{(l,r_d)} =\\ \text{softmax}_j \left(\mathbf{a}_{r_d}^{(l)T} \text{LeakyReLU}( \mathbf{W}_{r_d}^{(l)} \cdot [ \mathbf{h}_i^{(l)} \parallel \mathbf{h}_j^{(l)} ]) \right),
\end{equation}
where $r_d \in\mathcal{R}_d$ represents the set of edge types connecting nodes of different types. Messages are then computed as:
\begin{equation}
m_{j \rightarrow i}^{(l, r_d)} = \alpha_{jk}^{(l,r_d)} \mathbf{W}_{\phi(j),r_d}^{(l)} \mathbf{h}_j^{(l)}, \forall r_d \in \mathcal{R}_{\text{d}}, \phi(j) \neq \phi(k),
\end{equation}
\textbf{Aggregation and update:} After aggregating messages of both same-type and different-types, the node representations are updated as follows:
\begin{equation}
\mathbf{h}_{\phi(i)}^{(l+1)} = \text{SiLU} \left( \sum_{r \in \mathcal{R}_s \cup \mathcal{R}_d} \sum_{j \in \mathcal{N}_{r}(i)} m_{j \rightarrow i}^{(l, r)} \right).
\end{equation}

\subsection{Load Prediction}
\label{sec:htgnn_load_prediction}
Having extracted the context-aware dynamics of each node, we now combine the heterogeneous node representations to learn the virtual sensor function $f(\mathbf{X_T}^{t_l:t}, \mathbf{X_V}^{t_l:t}, \mathbf{W_T}^{t_l:t})= \mathbf{Y}^{t}$. 
We achieve this by flattening the final node representations into a unified input vector for a Multilayer Perceptron (MLP). The MLP processes this aggregated information and outputs two values: the predicted axial and radial loads.

To ensure the model's accuracy under real-world conditions, the training objective is to minimize the L1 loss between the predicted bearing load $\hat{\mathbf{y}}^{i}$ and the actual load ${\mathbf{y}}^{i}$. We choose L1 loss for its robustness to outliers. 
This is particularly important in bearing systems, occasional measurement noise or transitional operating conditions might generate extreme data points.
The loss is defined as
$\mathcal{L} = \frac{1}{M} \sum_{i=1}^{M} \left| \hat{\mathbf{y}}^{i} - \mathbf{y}^{i} \right|$,
where $M$ is the number of training samples.

\section{Case Study}
\label{sec:casestudy}
\begin{figure}[bthp]
  \centering
    \includegraphics[width=.85\linewidth]{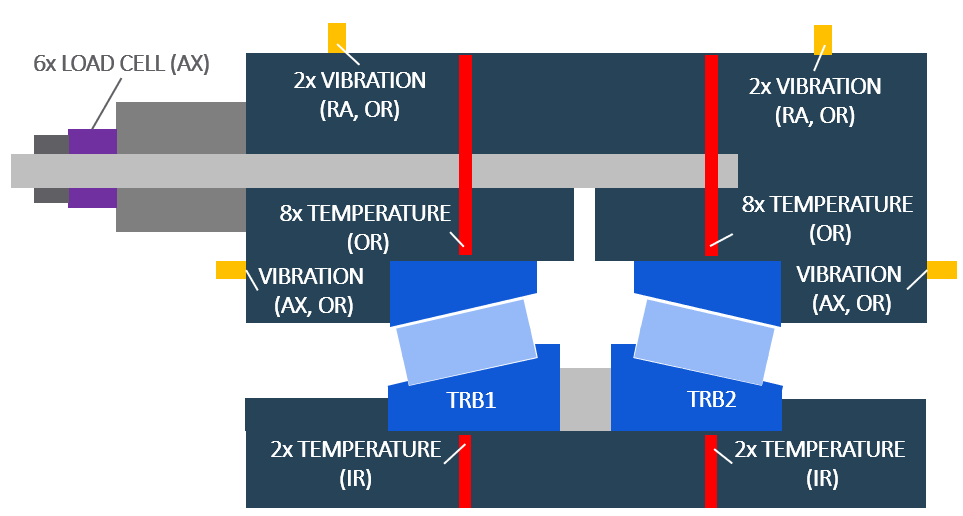}
  \caption{Test-rig configuration.}
  \label{fig:testrig_setup}
\end{figure}
The data used in this study was collected at the SKF Sven Wingquist Test Centre (SWTC) using a face-to-face test rig with two identical single-row tapered roller bearings (TRBs). The TRBs feature a rotating inner ring, an outer diameter of 2,000 mm, an inner diameter of 1,500 mm, and a width of 220 mm, each incorporating 50 rollers. 
This setup aims to assess load conditions under various operational scenarios.
Fig.~\ref{fig:testrig_setup} illustrates the sensor positioning on both identical TRBs. Ten temperature sensors are positioned on each bearing (eight uniformly distributed on the outer ring (OR), two on the inner ring (IR)). Additionally, six vibration sensors on the outer ring measure both axial (AX) and radial (RA) vibrations, with sensors placed at the top and bottom of the bearing housing for the radial direction. 

Temperature is recorded at a 1 Hz sampling rate with a precision of 0.05°C. Vibration data is resampled to 1 Hz through RMS aggregations. Axial and radial forces are measured and controlled by several load cells, with an aggregated load value in both directions used as a ground truth for this study (note that the radial load cells are not shown in the figure).

\subsection{Data Preprocessing}
To reduce noise and transient fluctuations in the temperature data, we apply a moving average filter with a 1-minute window. We focus on the rate of temperature change because the bearing temperature responds gradually to changes in load and speed. We calculate this rate over 5-minute periods to align with typical operational changes. This approach allows our model to identify the immediate impact of load changes on temperature, rather than the cumulative effects of historical variations.
After preprocessing, we split both temperature and vibration signals using a sliding window, with a length of 30 seconds and a stride of 1 second.

\subsection{Train-Test Split}
\begin{figure}[tbhp]
    \includegraphics[width=\linewidth]{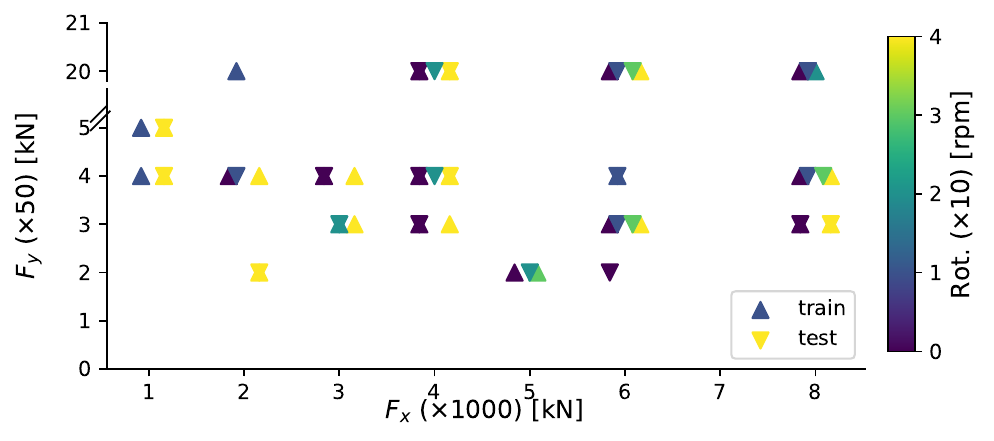}
  \caption{Train-test split of bearing load conditions for vibration data analysis (55\% training, 45\% testing)}
  \label{fig:split}
\end{figure}
To evaluate the model's performance on unseen data, we divided the dataset into training, validation, and testing sets. Approximately 55\% of the data (924,230 samples across 31 unique operating conditions) was allocated for training and validation, with a random 80/20 split. The remaining 45\% (699,340 samples across 25 unique operating conditions) was reserved for testing. We included only cases that maintained stationary operation for at least 10 minutes and up to 2 hours. 
Importantly, we define an operating condition (or case) as a unique combination of axial load $F_x$, radial load $F_y$, and rotational speed. 
In total, the dataset comprised 56 unique operating conditions. To assess generalization, 12 conditions present in the test set were excluded from the training and validation data. Fig.~\ref{fig:split} provides a detailed breakdown of the specific conditions included in the training and test sets.

\subsection{Heterogeneous Bearing Graph Construction}
\begin{figure}[tbhp]
  \centering
    \includegraphics[width=1.\linewidth]{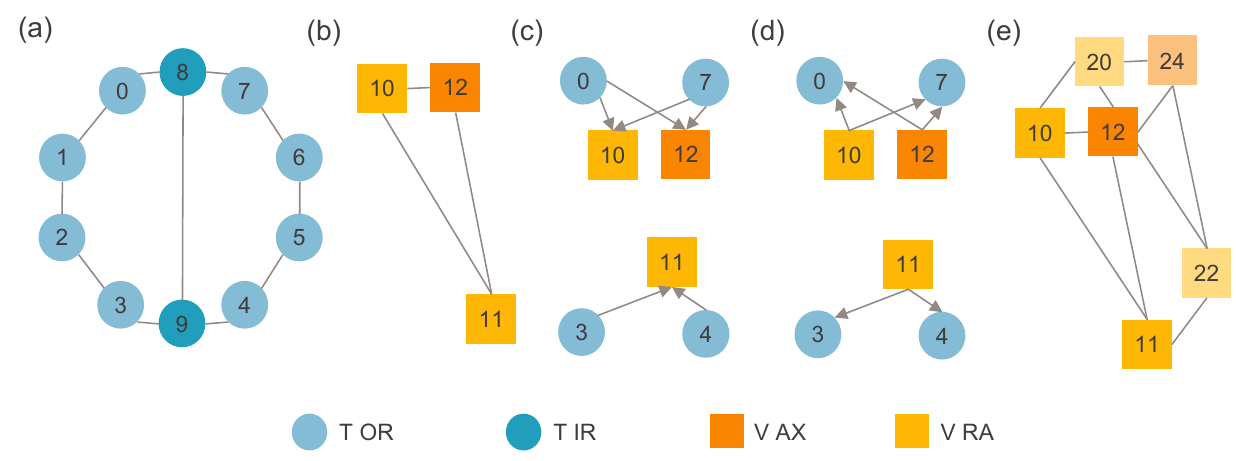}
  \caption{Heterogeneous graphs for bearing sensor network relationship modeling. (a) T-T (b) V-V (c) T-V (d) V-T (e) connectivity across two test rig bearings.}
  \label{fig:bearing_homogenous_graphs}
\end{figure}
We construct a heterogeneous graph with nodes representing sensors (temperature (T) and vibration (V)). Temperature nodes are further classified into inner ring (T IR) or outer ring (T OR) nodes. V nodes, which are installed on the outer ring, are distinguished by their load direction: radial (V RA) or axial (V AX). We model four types of relationships: T-T, V-V, T-V, and V-T. Here, T-T and V-V represent homogeneous relationships, while V-T and T-V represent heterogeneous relationships. Node positions reflect physical sensor placement. Fig.~\ref{fig:bearing_homogenous_graphs}\textcolor{cyan}{(a)} and \textcolor{cyan}{(b)} illustrate the connectivity within a single bearing based on physical proximity. Additionally, IR nodes are connected due to relatively uniform temperatures across the inner ring. Given that the test rig consists of two bearings, we connect them based on proximity, as illustrated in Fig.~\ref{fig:bearing_homogenous_graphs}\textcolor{cyan}{(e)} for V nodes. We assume symmetrical (undirected) relationships within the same sensor type and model heterogeneous T-V and V-T relationships with directed edges, as demonstrated in Fig.~\ref{fig:bearing_homogenous_graphs}\textcolor{cyan}{(c)} and \textcolor{cyan}{(d)}.

\subsection{Experimental Setup}
\textbf{Baseline.} We employ a 1DCNN model as our baseline due to its established success in handling multivariate time series data. 1DCNNs are particularly well-suited for signal prediction tasks, making them a strong baseline.
We adapt the design from \shortcite{chao2022fusing}, tailoring the architecture to our specific dataset through a grid search to minimize the mean absolute error (MAE) on the validation set. 
The explored parameter spaces included hidden channel dimension (20, 50, or 100), kernel size (3, 5, or 9), number of channels (20, 50, or 100), and number of layers (3, 4, or 5). 
The optimized model consists of four layers, each with 100 channels with 100 hidden dimensions, a kernel size of 9, batch normalization, a dropout rate of 0.5 for regularization, and a SiLU activation function (consistent with our proposed method). This configuration has a total of 209,403 parameters.

\textbf{HTGNN hyperparameter tuning.}
We similarly used grid search for HTGNN hyperparameter tuning. 
To reduce the search space, we maintained a consistent hidden size across all layers and the same graph embedding dimension for all GNN modules.  
The search space comprised: node embedding dimension (values of 10, 15, 20), number of GNN layers (2 or 3), GNN hidden dimension (40 or 80), graph head hidden dimension (40 or 80), and number of graph head layers (2 or 3). The optimal HTGNN configuration consists of a node embedding dimension of 10, 3 GNN layers with a hidden dimension of 80, and a graph head dimension of 40. The configuration has a total of 142,394 parameters.

\textbf{Training.} We optimized the HTGNN and 1DCNN models using the AdamW optimizer with a learning rate of 1e-3. Training was continued for up to 50 epochs with early stopping at 30 epochs with patience of 10 steps. 
We used a batch size of 512 and minimized L1 loss (defined in Sec.~\ref{sec:htgnn_load_prediction}).
To ensure the robustness of our results, experiments were repeated five times with different initializations, and the mean and standard deviation of the results were reported. 

\section{Results}
\label{sec:result}
We evaluate the model performance on Mean Absolute Error (MAE) and Mean Absolute Percentage Error (MAPE).

\textbf{Superior performance on seen conditions.} 
Tab.~\ref{tab:model-performance} highlights the HTGNN model's superior performance advantage compared to traditional 1DCNN models in predicting seen conditions. Notably, this improvement is evident in both axial (Fx) and radial (Fy) load predictions, with the HTGNN achieving approximately one-third the MAPE for Fx and half the MAPE for Fy compared to the 1DCNN. 
Importantly, the scenario considered here reflects real-world conditions. It is feasible for the sensor roller to collect load data across all typical operating conditions before its battery depletes, allowing the HTGNN to serve as a reliable virtual sensor.

\textbf{HTGNN's physical prior.} The superiority of the HTGNN in unseen conditions highlights the advantages of explicitly modeling heterogeneous sensor relationships. 
The physical connectivity in the bearing system acts as an effective inductive bias for the model.
We attribute the improved performance of the HTGNN to its ability to capture complex interactions between temperature and vibration measurements, which often exhibit interdependent behaviors in bearing systems. The proposed architecture of the HTGNN is ideally suited to represent these heterogeneous relationships.
In contrast, 1DCNN's homogeneous approach to processing variables limits its ability to model such complex interdependencies, leading to higher prediction errors.

\begin{table}[btp]
\centering
\caption{Averaged model performance over cases and runs}
\label{tab:model-performance}
\begin{tabular}{crrr}
\toprule
 &  & \textbf{1DCNN} & \textbf{HTGNN} \\ \midrule
\multirow{4}{*}{Seen}   
     & MAE\textsubscript{\(F_x\)} (kN)  & 531.3  & \textbf{203.1}  \\ 
     & MAE\textsubscript{\(F_y\)} (kN)   & 33.2   & \textbf{12.4}   \\ 
     & MAPE\textsubscript{\(F_x\)} (\%) & 12.8  & \textbf{4.5}  \\ 
     & MAPE\textsubscript{\(F_y\)} (\%) & 12.0  & \textbf{5.7}  \\ \midrule
\multirow{4}{*}{Unseen} 
    & MAE\textsubscript{\(F_x\)} (kN)  & 1765.5 & \textbf{1649.7} \\ 
    & MAE\textsubscript{\(F_y\)} (kN)  & 58.7   & \textbf{57.4}   \\ 
    & MAPE\textsubscript{\(F_x\)} (\%) & 33.2  & \textbf{29.2}  \\ 
    & MAPE\textsubscript{\(F_y\)} (\%) & 17.8  & \textbf{15.8}  \\ 
\bottomrule
\end{tabular}
\end{table}
\begin{figure}[bthp]
  \centering
    \includegraphics[width=1\linewidth]{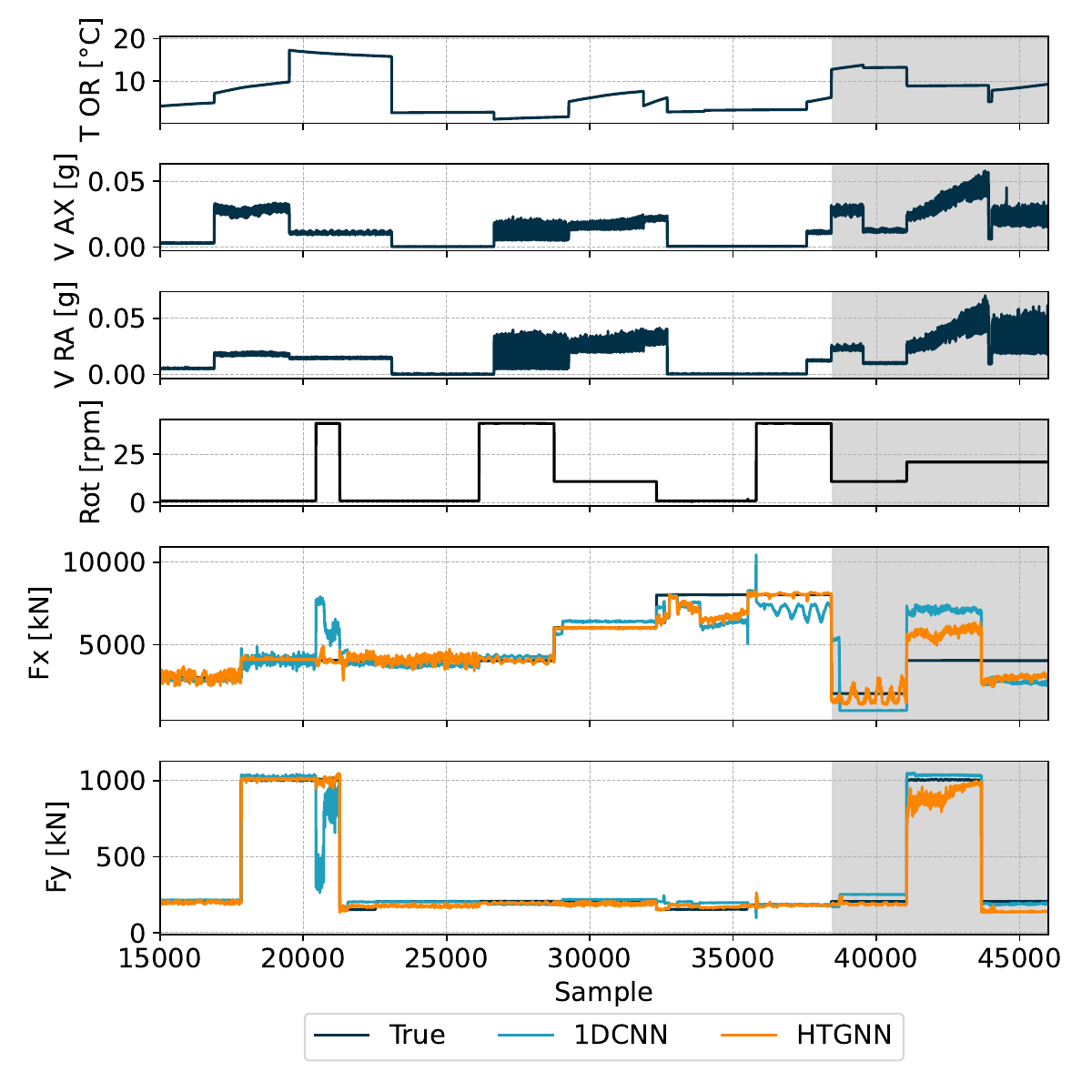}
  \caption{Examples of input signals and load prediction performance. Shaded areas indicate unseen conditions.}
  \label{fig:result_plot}
\end{figure}

\begin{figure*}[tbhp]
  \centering
    \includegraphics[width=.9\linewidth]{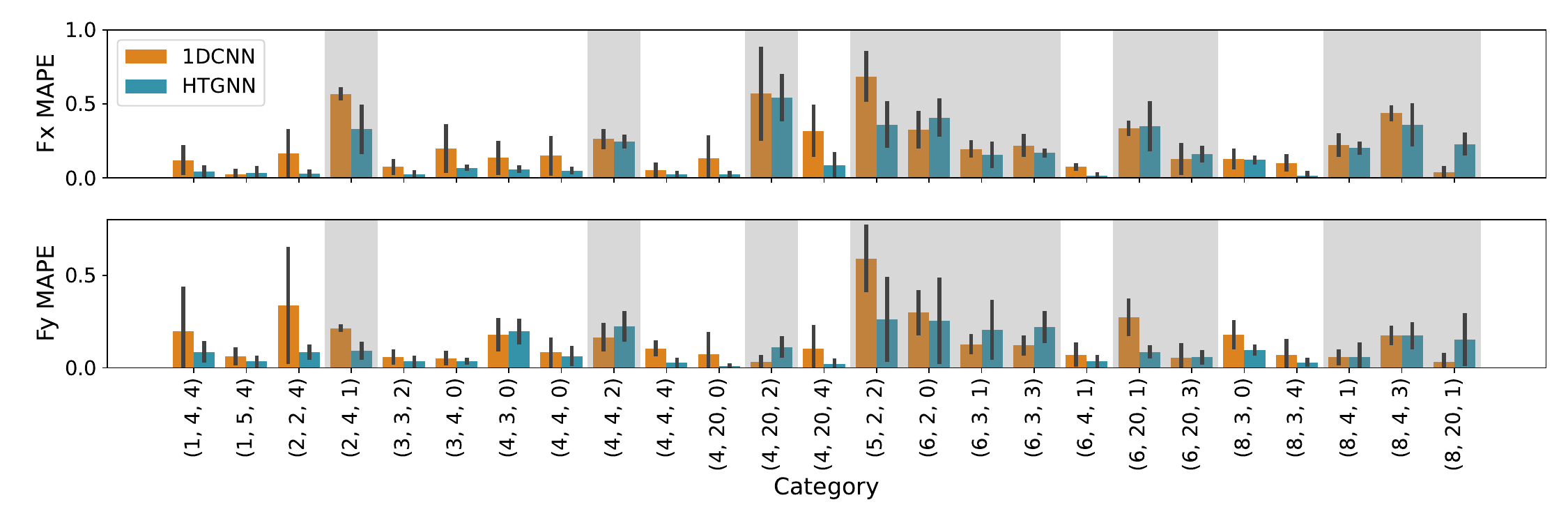}
  \caption{Mean test set performance averaged over 5 runs of CNN and HTGNN on bearing load conditions $F_x (\times 1000)$ [kN], $F_y (\times 50)$ [kN], and rotational speed ($\times 10$) [r/min]. Shaded areas indicate unseen conditions.}
  \label{fig:result_by_case}
\end{figure*}
\textbf{Better generalizability.} 
Fig.~\ref{fig:result_by_case} presents the mean MAPE in $F_x$ and $F_y$ for various bearing load conditions, with unseen conditions highlighted in gray. Although the HTGNN generally outperforms the CNN in handling unseen conditions, as detailed in Tab.~\ref{tab:model-performance}, there are instances depicted in Fig.~\ref{fig:result_by_case} where CNN shows competitive performance.
This challenge in generalization can be partially attributed to the dynamics shown in Fig.~\ref{fig:result_plot}, which illustrates the significant effects of rotational speed changes on both vibration intensity and the rate of temperature change. 
Additionally, the underrepresentation of certain rotational speeds in the training data may impede interpolation, impacting the generalization capabilities of both models. Interestingly, as depicted in Fig.~\ref{fig:result_plot}, the HTGNN tends to generalize better for the commonly encountered unseen rotational speed of 10 rpm than for the less frequently occurring speed of 20 rpm, across both axial and radial loads. For details on the distribution of conditions in the training and testing sets, see Fig.~\ref{fig:split}.

\section{Conclusion}
\label{sec:conclusion}
In this research, we propose HTGNN, a novel virtual sensor that accurately maps vibration and temperature signals under varying rotational speeds to axial and radial bearing load predictions. 
Our findings demonstrate that HTGNN outperforms 1DCNN models, particularly when trained on representative conditions. 
The success of HTGNN highlights the importance of incorporating physical priors and inductive biases: by modeling the connectivity of the bearing sensor network, HTGNN effectively captures the complex interactions between temperature and vibration. This superior performance suggests HTGNN's potential as a reliable virtual sensor in real-world applications, replacing battery-powered load sensors after their lifespan.
This could facilitate proactive maintenance, reducing unexpected breakdowns and optimizing the lifespan of bearings.
However, the models cannot generalize as effectively to unseen speed conditions. Future work should focus on investigating datasets that include a broader range of speed conditions in the training to determine if the model can improve its generalization capabilities. Additionally, measuring the model's performance using real load data measured from sensor rollers in real operations and not just from the test rig would be valuable.

\bibliographystyle{apacite}
\PHMbibliography{ijphm}

\end{document}